\documentclass{article}
\usepackage{spconf,amsmath,graphicx}
\usepackage{subfigure}
\usepackage{multicol}  
\usepackage{multirow}
\usepackage{pifont}
\usepackage{wrapfig} 
\usepackage{booktabs}  
\usepackage{threeparttable}  
\usepackage[colorlinks,
            linkcolor=blue,
            anchorcolor=blue,
            citecolor=blue]{hyperref}   
\usepackage[numbers,sort&compress]{natbib} 

\usepackage{bm}

\def \xx {\bm{x}}
\def \ff {\bm{f}}
\title{Adaptive Frequency Learning in Two-branch Face Forgery Detection}
\name{Neng Wang$^{\dagger,\amalg}$, Yang Bai$^{\star,\amalg}$\thanks{ $^\amalg$Equal contributions}, Kun Yu$^{\dagger}$, Yong Jiang$^{\star}$, Shu-tao Xia$^{\star,\ddagger}$\thanks{ $^\ddagger$Corresponding author: Shu-tao Xia}, Yan Wang$^{\dagger}$
\thanks{This research is supported in part by the National Key Research and Development Program of China under Grant 2020AAA0140000.}}
\address{$^{\star}$ Tsinghua University, China \\
  $^{\dagger}$ Alibaba Security AI Lab, China  \\
  \tt \small linqing.wn@alibaba-inc.com, y-bai17@mails.tsinghua.edu.cn}

\begin{document}
\maketitle

\begin{abstract}
Face forgery has attracted increasing attention in recent applications of computer vision. Existing detection techniques using the two-branch framework benefit a lot from a frequency perspective, yet are restricted by their fixed frequency decomposition and transform. In this paper, we propose to \textbf{A}daptively learn \textbf{F}requency information in the two-branch \textbf{D}etection framework, dubbed \textbf{AFD}.
To be specific, we automatically learn decomposition in the frequency domain by introducing heterogeneity constraints, and propose an attention-based module to adaptively incorporate frequency features into spatial clues. Then we liberate our network from the fixed frequency transforms, and achieve better performance with our data- and task-dependent transform layers. Extensive experiments show that AFD generally outperforms.
\end{abstract}
\section{Introduction}
In recent years, deep learning techniques have been widely deployed in various applications and achieved remarkable success \cite{larochelle2009exploring}.
However, the development of deep-learning-driven application is still at the risk of AI security issues,
among which the face forgery is specially studied for its broad and important applications such as in facial payment or social media.
Nowadays, such face forgery has been  specially studied with DeepFake techniques, which conceal the forgery artifacts by extracting and modifying latent characteristics from the original images.


To defend against manipulations, various detection techniques have been proposed. Some are based on time-series clues \cite{masi2020two}, undermining frame-to-frame inconsistencies and resulting in high computational costs. While some are based on frame-level clues \cite{wang2020cnn}, dealing with frame-level (image) information, among which some only works in the spatial domain and some considers the frequency information \cite{qian2020thinking}, thus developing a two-branch detection framework. 
The two-branch framework outperforms due to two perspectives. From one aspect, recent forgery methods mainly rely on deep learning techniques with convolution layers, which could leave some periodic clues in the frequency domain \cite{wang2020cnn}.
From the other aspect, easily applied frequency decomposition could help describe differently frequency-conscious patterns, benefiting the feature extraction capacity only in spatial domain.
However, existing two-branch detection techniques usually either adopt a simple frequency decomposition or fixed frequency transforms. Moreover, they are often in lack of feature fusion exploration incorporating two branches, showing potential bottlenecks. 

Inspired by the analyses and studies above, learning in an adaptive manner in frequency domain to extract more sufficient information is encouraged.
In this paper, we propose to \textbf{A}daptively learn \textbf{F}requency information in the two-branch \textbf{D}etection framework, dubbed \textbf{AFD}. 
Our key contributions are summarized as follows:
\begin{itemize}
    \item We propose to adaptively decompose the frequency information with soft masks optimized using triplet loss, and replace the fixed frequency transforms with fine-tuned learnable parameters.
    \item We propose to incorporate frequency characteristics into spatial clues with an attention module, making better feature fusion in the two branchs. 
    \item Extensive experiments show the consistent superiority of our proposed AFD especially in the two-branch framework. Besides, sufficient ablation studies provide a detailed and in-depth understanding of AFD detection.
\end{itemize}

\section{Related Works}
Face forgery with deep learning skills usually works in the fashion of manipulating original images on the extracted characteristics, \textit{e.g.}, DeepFake, FaceSwap, Face2Face and NeuralTextures, etc. In addition, GANs are also widely deployed\cite{xuan2019generalization} in the generation of forged images without any original image specifically in pairs. 
While forgery techniques are showing increasingly improved visual performance, the detection techniques are also thriving.

Since a large number of forgeries are towards videos, these detection techniques can be categorized into time- and frame-based ones.
\textit{Time-based detection} techniques capture temporary inconsistencies between the frames, often on scenarios of emotion or background, etc. Besides using CNNs to extract frame-level features, they usually utilize RNNs as time-aware pipelines to learn manipulation clues across frames \cite{masi2020two}. 
Due to their high computational cost on video-level, recent studies also explore \textit{frame-based detection}, processing the videos into frame-wise images and extracting frame-level characteristics. 

\noindent
\textbf{Spatial only} MesoInception-4 \cite{afchar2018mesonet} is a CNN-based network to detect forged videos. GAN Fingerprint \cite{yu2019attributing} discovers the specific manipulation patterns introduced by different GANs. 

\noindent
\textbf{Frequency only}
Besides spatial-based detection methods, the frequency-based detection methods are proposed to capture features in the frequency domain. As recent manipulation techniques often adopt convolution layers, the manipulated images are more likely to leave obvious clues and present certain periodic signals in the frequency domain \cite{wang2020cnn}. 
\cite{liuspatial} proposes to do the classification using phase information in the frequency domain with DFT\cite{smith2007mathematics}. \cite{chen2017jpeg} uses high-pass filters, \textit{e.g.}, Gabor filters, to extract edge and texture information regarding high-frequency components.

\noindent
\textbf{Two-branch}
In addition, several techniques have explored the frequency clues under the two-branch framework. 
\cite{durall2019unmasking} proposes to learn the frequency information by averaging the amplitudes of the different frequency bands with Discrete Fourier Transform (DFT). 
\cite{qian2020thinking} proposes to learn more about features in the broader frequency bands by decomposing the frequency information.
However, all of the above two-branch techniques deploy the frequency decomposition independent from spatial domain (RGB branch) and a fixed frequency transform fashion as well, which are the main bottlenecks. In this paper, we thus explore the adaptive frequency learning under two-branch framework.


\section{Proposed Methodology}

\noindent
\textbf{Discrete Cosine Transform}
Amongst the widely applied frequency transform techniques, Discrete Cosine Transform (DCT) \cite{khayam2003discrete} and its inverse (IDCT) are the most common ones.
In this paper, $\mathcal{D}/\mathcal{D^{-\text{1}}}$ indicate applying DCT/IDCT. $S_{\mathcal{D}}/S_{\mathcal{D^{-\text{1}}}}$ indicate signals after the DCT/IDCT operation.

Given data from $k$ different manipulation methods (\textit{i.e.}, $k$ different domains or data sets), the $i$-th data set has $N_{i}$ images, $F$ and $T$ indicate the main feature extractor and classifier in the detection framework, then the original cross-entropy objective function is defined as follows,
\begin{equation}
    \mathcal{L}_{\text{ce}} = \sum_{i=1}^{k} \sum_{j=1}^{N_{i}} -y_{j}^{i} \cdot \log T(F(\xx_{j}^{i})),
\end{equation}
where $(\xx_{j}^{i},y_{j}^{i})$ indicates the $j$-th image and label data pair from the $i$-th domain.

\begin{figure}[t]
\vspace{-0.3cm}
\includegraphics[width=9.0cm]{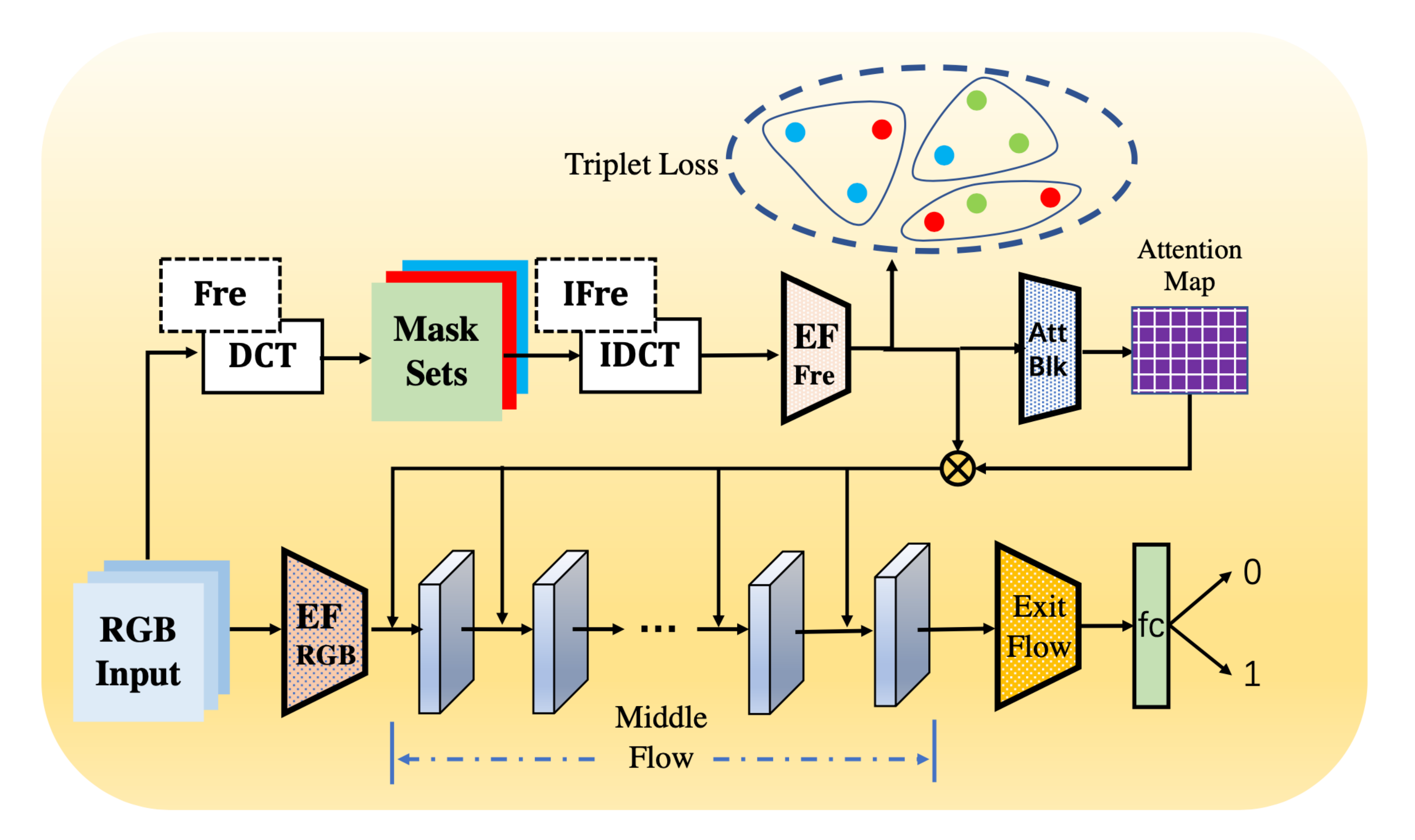}
\centering
\vspace{-0.5cm}
\caption{The pipeline of our AFD under two-branch detection framework. We use three splits. `Soft Masks' are multiplied on $S_{\mathcal{D}}$ and followed with IDCT. The features extracted by `EF Fre' from frequency branch (`F Fre') are fused with original RGB branch (`F RGB') using an attention module (`Att Blk'). `Triplet loss' and `Softmax operation' are introduced in optimizing soft masks. `Middle/Exit Flow' indicate different layers of Xception. Then original DCT/IDCT is replaced with fine-tuned `Fre'/`IFre' layers.}
\label{fig:AdaD}
\end{figure}

To sufficiently extract characteristics in frequency branch, the decomposition is widely applied yet either using fixed binary masks on $S_{\mathcal{D}}$ or rarely incorporating with the RGB branch, which restricts the detection performance.
Motivated by this, we propose to adaptively decompose the frequency information by optimizing soft masks. During the optimization of soft masks, the fusion of decomposed frequency-branch features with the original RGB branch is also considered to improve the two-branch detection efficiency. Then the adaptive frequency transform is deployed. The overall details of our AFD are shown in Fig. \ref{fig:AdaD}.

\subsection{AdaD: Adaptive Frequency Decomposition}
\label{sec:AdaD}
We propose to decompose the frequency bands by applying and optimizing soft masks. Given the size of an input image as $w \times h$, the number of splits or masks on $S_{\mathcal{D}}$ as $n$, we initialize soft masks $M_i (i=1,2,...,n)$ in two ways, using 1) given binary masks to benefit from stronger priors, which are linear-wise splits with given hyper-parameters due to a strong `energy compaction' property of DCT, or 2) average masks with all elements as $1/n$. To encourage the differently frequency-conscious pattern, we introduce triplet loss \cite{deng2020rethinking} as a heterogeneity constraint, which is defined as
\begin{equation}
\begin{aligned}
&\ff_{i} = F_{\text{fre}}(\mathcal{D^{-\text{1}}}(S_{\mathcal{D}} \times M_i)),\\
&\mathcal{L}_{\text{trip}} = \frac{1}{N} \sum_{(M_a,M_p,M_n)} \max(\|\ff_{a}-\ff_{p}\|^2-\|\ff_{a}-\ff_{n}\|^2 + m, 0),
\end{aligned}
\label{eq:trip}
\end{equation}
where $m$ is the margin, $M_a$ indicates an anchor mask, $M_p$ is the same as $M_a$, while $M_n$ is chosen from the left $N-1$ masks in turn. $\ff$ represents for the correspondingly extracted feature.

\begin{wrapfigure}{l}{5.5cm}
\vspace{-0.4cm}
\subfigure[hard masks]{
\label{fig:1a}
\includegraphics[width=5.5cm]{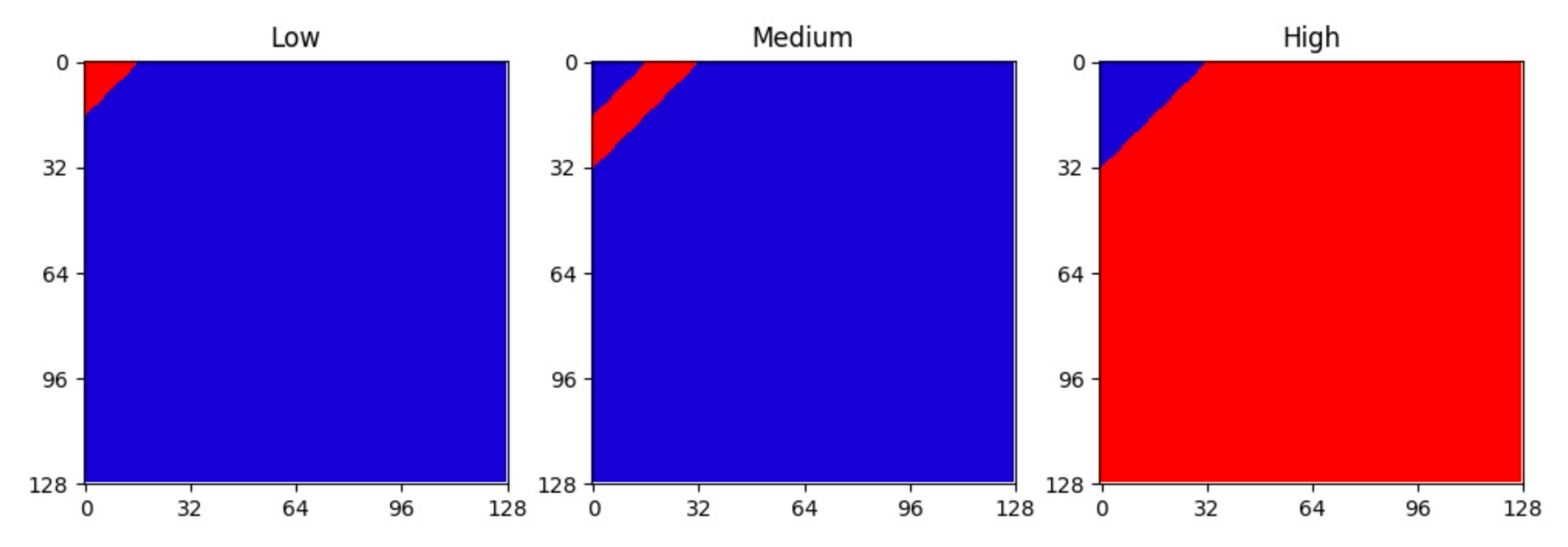}}
\vspace{-0.2cm}
\subfigure[soft masks]{
\label{fig:1b}
\includegraphics[width=5.5cm]{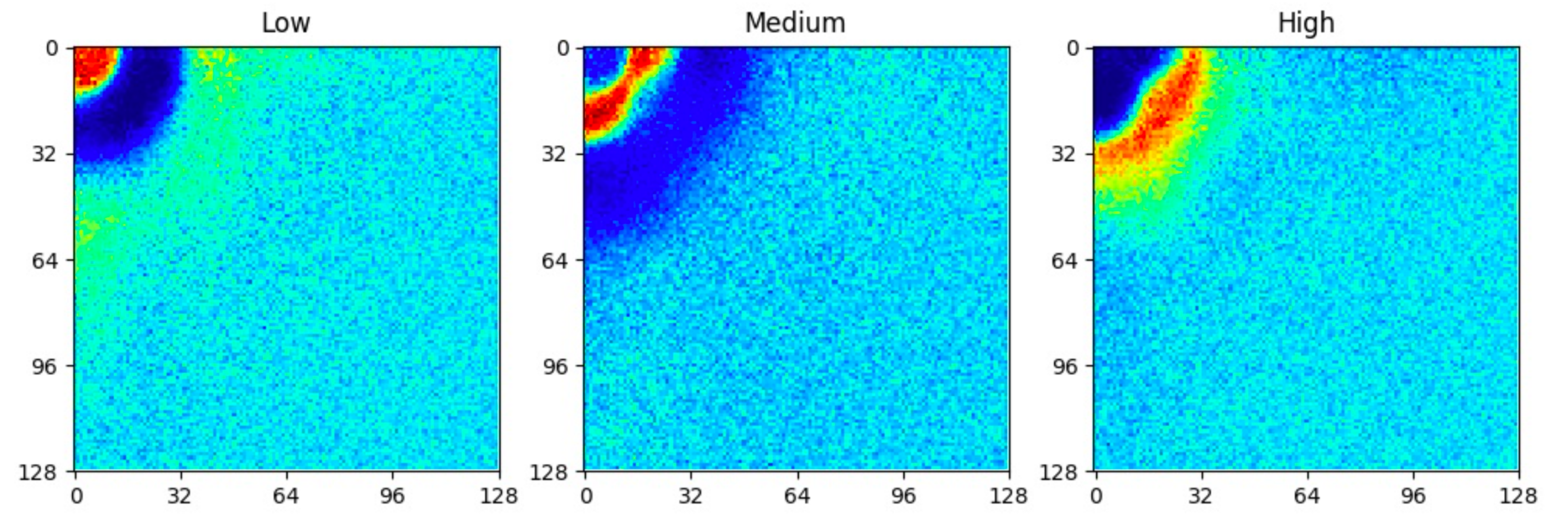}}
\caption{(a) Hard and (b) optimized soft masks in frequency decomposition.}
\vspace{-0.5cm}
\label{fig:softmask}
\end{wrapfigure}
Furthermore, we apply softmax normalization across the $n$ masks to make sure all information in the frequency domain are preserved. The optimized soft masks by our method are shown in Fig. \ref{fig:1b}, which are clearly more data- and task-dependent compared with the initial hard masks in Fig. \ref{fig:1a}.

\subsection{EFF: Exploring Feature Fusion}
The frequency-decomposed images with soft masks, $\mathcal{D^{-\text{1}}}(S_{\mathcal{D}})$, are fed into a shared feature extractor $F_{\text{fre}}$. Then we fuse their features with the original RGB branch using an attention module \cite{vaswani2017attention}, which is mainly composed of linear layers with ReLU activation. 
For a network with $n$ soft masks and $m$ candidate entrance layers in the RGB branch backbone, the attention map $Att$ is designed with the size of $n \times m$ and softmax normalized along the layers, then it element-wise re-weights the extracted frequency components for feature fusion, following 
\begin{equation}
    F = F_{\text{RGB}} + Att \times F_{\text{fre}}.
\end{equation}
It is important to denote that $Att$ requires no supervision, which is automatically learnt during the training phase.
Then the objective function in AdaD is combined with cross-entropy and triplet losses as
\begin{equation}
    \mathcal{L}=\gamma \cdot \mathcal{L}_{\text{trip}} + \mathcal{L}_{\text{ce}}.
\label{eq:AdaD}
\end{equation}

\subsection{AdaT: Adaptive Frequency Transform}

Existing detection methods often apply the fixed frequency transform, decreasing detection performance.
In this section, we model the transforms as learnable layers, and achieve significant improvement in a simple yet effective fine-tuning manner.
As shown in Fig. \ref{fig:AdaD}, we first train the detection framework applying AdaD. Next, we replace the DCT/IDCT operation with the learnable pre-processing `Fre'/`IFre' layers, which are initialized with the corresponding DCT/IDCT parameters and then fine-tuned. In such a fashion, the pre-processing `Fre'/`IFre' layers could estimate and improve the standard DCT/IDCT. 

\section{Experiments}
\subsection{Experimental Setups}
\vspace{0.1cm}
\noindent
\textbf{Datasets}
We conduct our experiments on FaceForensics++ \cite{rossler2019faceforensics++}, which is a face forgery video data set containing 1,000 real videos (720/140/140 for training/validation/test).
Each video undergoes 4 manipulation methods. We extract 50 frames uniformly from each video in C40 data set with low quality (heavy compression). Then we crop the facial images and resize them into $299\times299\times3$ using MTCNN \footnote{https://github.com/timesler/facenet-pytorch}. In the training/test phase, to ease the data imbalance issue, we re-sample original images and mix them with each manipulated images separately. We use the test sets of all four manipulated data and original data as the \textit{whole test set}, while we separately mix the manipulated data and original data as the \textit{domain-specific test set}.


\begin{table}[t]
  \caption{Evaluations on the detection effectiveness on whole test sets. `Xception $\star$' indicates our backbone. 
  The best results are \textbf{boldfaced}.}
  \label{tab:acc1}
  \centering
  \begin{tabular}{c|cc}
    \toprule
    Methods & ACC (\%) & AUC (\%)\\
    \midrule
    Steg. Features \cite{fridrich2012rich}& 55.98 & -- \\
    LD-CNN \cite{cozzolino2017recasting} & 58.69 & --\\
    Constrained Conv \cite{bayar2016deep} & 66.84 & --\\
    CustomPooling CNN \cite{rahmouni2017distinguishing} & 61.18 & --\\
    MesoNet \cite{afchar2018mesonet} & 70.47 & --\\
    Face X-ray \cite{li2020face} &  -- & 61.60 \\
    Xception \cite{chollet2017xception} & 81.00 & --\\
    Two-branch \cite{masi2020two} & 86.34 & 91.10\\
    \midrule
    Xception $\star$ & 89.39 & 91.75 \\
    AFD (ours) $\star$ & \textbf{90.33} & \textbf{94.24}\\
    \bottomrule
  \end{tabular}
\vspace{-0.4cm}
\end{table}

\begin{table}[!h]\centering
  \caption{Evaluations on the effectiveness of our detection methods on domain-specific test sets. Our method outperforms. The best results are \textbf{boldfaced}.}
  \label{tab:acc2}
  \centering
  \begin{tabular}{c|c|cc}
    \toprule
    Data sets & Methods & ACC (\%) & AUC (\%) \\
    \midrule
     \multirow{3}{*}{Face2Face} & Xception $\star$ & 80.95 & 89.35\\
    & AFD (ours) $\star$ & \textbf{82.00} & \textbf{91.99} \\
    \cline{1-4}
     \multirow{3}{*}{FaceSwap} & Xception $\star$ & 82.09 & 91.61 \\
    & AFD (ours) $\star$ & \textbf{82.45} & \textbf{92.90} \\
    \cline{1-4}
     \multirow{3}{*}{NeuralTexture} & Xception $\star$ & 84.20 & 94.22\\
    & AFD (ours) $\star$ & \textbf{85.71} & \textbf{97.08}\\
    \cline{1-4}
     \multirow{3}{*}{DeepFakes} & Xception $\star$ & 82.37 & 91.83\\
    & AFD (ours) $\star$ & \textbf{84.30} & \textbf{94.99}\\
    \bottomrule
  \end{tabular}
  \vspace{-0.3cm}
\end{table}

\begin{table}[h]
\caption{Ablation study on different decomposition settings in frequency branch. Overall, the models using soft masks with binary initial and optimized by triplet loss perform better. The best results are \textbf{boldfaced}.}
\label{tab:AdaD_abl2}
\centering
\begin{tabular}{ccc|cc}
    \toprule 
    Mask & Init & Optimization & ACC(\%) & AUC(\%)\\
    \midrule
    Hard & Binary & -- & 87.93 & 91.86 \\
    \midrule
    Soft & Average & -- & 88.57 & 91.93 \\
    Soft & Average & Triplet & 89.42 & 92.48\\
    Soft & Average & Softmax+Triplet & 88.56 & 92.81\\
    \midrule
    Soft & Binary & -- & 89.25 & 92.58\\
    Soft & Binary & Triplet & \textbf{89.93} & 92.69\\
    Soft & Binary & Softmax+Triplet & 89.61 & \textbf{93.17} \\
    \bottomrule
\end{tabular}
\vspace{-0.5cm}
\end{table}

\vspace{0.1cm}
\noindent
\textbf{Implementation Details}
We use a pre-trained Xception\footnote{https://github.com/Cadene/pretrained-models.pytorch} as our backbone. The networks are then optimized with Adam optimizer \cite{zhang2018improved}, with initial learning rate set as 0.001. We use cosine annealing scheduler \cite{loshchilov10sgdr} to adjust our learning. The batch size is set as 32. In AdaD, we train for 15 epochs using $\gamma = 1.0$ (Eq. \ref{eq:AdaD}) and $m = 0.1$ (Eq. \ref{eq:trip}). The binary masks are initialized as 32/64 in the decomposition of low/medium/high frequency bands. In AdaT, we fine-tune for 2,500 iterations.


\subsection{Experimental Results}
\label{sec:exp1}





We first compare the detection performance with other baseline methods in Tab. \ref{tab:acc1}. 
As for the baseline methods, we apply the test results from the recent paper \cite{qian2020thinking,zhaomulti}. For a more fair and convincing comparison, we also train an Xception backbone method on the our extracted data with our implementation settings, which is specifically marked with $\star$ in all tables. 
The AUC is improved from 91.75\% to 94.24\% compared with baseline Xception $\star$ backbone, demonstrating the superiority of our methods. 

In Tab. \ref{tab:acc2}, we further evaluate the detection results on domain-specific test sets. Overall, our method outperforms on all test sets,
\textit{e.g.}, the AUC arises from 91.83\% to 94.99\% when evaluated on DeepFakes data set. It is shown that our proposed models have achieved the consistent improvement and represented the best results when evaluated on both whole test set or domain-specific test sets in Tab. \ref{tab:acc1} and Tab. \ref{tab:acc2}. Therefore, the evaluations in this section show the superiority of our methods in the detection effectiveness.

\subsection{Ablation Study}
We explore different tricks of frequency decomposition in Tab. \ref{tab:AdaD_abl2} and feature fusion in Tab. \ref{tab:eff}. In this section, we take three splits as an example. 

As for the initial masks, we use 1) average masks whose elements are all 1/3 (\textit{i.e.}, `Average Init') and 2) binary masks which are linearly splitted by 32/64 as in Fig. \ref{fig:1a} (\textit{i.e.}, `Binary Init').
We also analyse on triplet loss and softmax operation in optimizing soft masks to make sure all information in the frequency domain are better preserved.
Experimental results show that soft masks initialed with the binary masks show overall better performances than the average ones. In addition, triplet loss and softmax operation can generally improve the detection performances regardless of the initial masks. 
For example, evolved from the models applying soft masks with binary initial, triplet loss and softmax operation, the AUC could be improved to 93.17\% from baseline 91.86\%. 

Besides the attention module we used in our proposed AFD to adaptively fuse features in two branches, we also deploy the feature fusion operation at the entry/exit of middle flow in Xception as the ablation study. In addition, we design one predefined fusion fashion to encourage low-frequency features to be fused at entry layer of Xception while the high-frequency ones at the exit. Results in Tab. \ref{tab:eff} show that the adaptive attention module adpoted in our AFD outperforms.

\vspace{-0.5cm}
\begin{table}[h]
\caption{Ablation study on feature fusion settings for two branches. The attention-based module performs best. The best results are \textbf{boldfaced}.}
\label{tab:eff}
\centering
\begin{tabular}{c|cc}
    \toprule 
    Fusion Methods & ACC(\%) & AUC(\%)\\
    \midrule
    All At Entry & 88.57 & 91.49\\
    All At Exit & 89.19 & 92.86\\
    Predefined & 89.11 & 92.31\\
    Attention-based & \textbf{89.61} & \textbf{93.17} \\
    \bottomrule
\end{tabular}
\vspace{-0.5cm}
\end{table}

\section{Conclusions}
In this paper, we propose AFD, to encourage the adaptive frequency learning in two-branch forgery detection framework.
We adaptively learn the frequency decomposition by optimizing soft masks in AdaD, conducting feature fusion of two branches with an attention module in EFF and replace the fixed frequency transform with adaptive AdaT in a fine-tuning fashion.
Extensive experimental results demonstrate the consistent superiority of our proposed AFD. The detailed ablation studies provide in-depth analyses on the essence of adaptive frequency learning in detection tasks.

\bibliographystyle{IEEEbib}
\bibliography{refs}

\appendix
\end{document}